\documentclass[runningheads]{llncs}
\usepackage[T1]{fontenc}
\usepackage{graphicx}
\usepackage{amsmath, amssymb}
\usepackage{subcaption}
\usepackage{url}
\usepackage{xurl}

\usepackage[normalem]{ulem}
\usepackage{comment}
\usepackage{multirow}
\usepackage{ulem}

\usepackage[table,xcdraw]{xcolor}
\usepackage{booktabs, cellspace}

\begin{document}
\title{Semantic Audio-driven Understanding for Dynamic Humanoid Whole Body Control}
%
%
\author{
J. M. A. Marcelo \inst{1} 
\orcidID{0009-0005-7413-6618}\and
M. Brienza \inst{1} 
\orcidID{0009-0000-1549-9500}\and
E. Bugli \inst{1} 
\orcidID{0009-0000-9540-681X}
\and
L. Comito \inst{1} 
\orcidID{0009-0000-3827-0581}
\and
D. Nardi\inst{1}\orcidID{0000-0001-6606-200X}
\and
D. D. Bloisi \inst{2}\orcidID{0000-0003-0339-8651}
\and
\\V. Suriani \inst{1}  
\orcidID{0000-0003-1199-8358}
}
\authorrunning{Marcelo et al.}
%
\titlerunning{Semantic Audio for Dynamic Humanoid Control}
\institute{Dept. of Computer, Control, and Management Engineering\\ Sapienza University of Rome, Rome (Italy),
    \email{\{lastname\}@diag.uniroma1.it} \and
Dept. of International Humanities and Social Sciences, 
International University of Rome, Rome (Italy),
\email{domenico.bloisi@unint.eu}
}
\maketitle             

\begin{abstract}
Recent advances in humanoid robotics and reinforcement learning have enabled the acquisition of highly expressive whole-body motion policies. However, most robotic performances remain based on pre-scripted sequences or externally triggered behaviors, limiting autonomy and responsiveness to dynamic environments. In this work, we introduce a novel multi-modal orchestration framework for semantic audio-driven humanoid control, enabling robots to autonomously select and execute appropriate motion skills in real time. The system processes continuous audio streams and routes them into music or speech branches. Music input is handled via audio fingerprinting and semantic embeddings to retrieve track identity and temporal alignment, allowing dynamic mapping between musical segments and motion policies. Speech input is grounded into a discrete library of imitation-learned skills, enabling direct human-robot interaction. Both modalities share a unified interface that schedules skill execution over a reinforcement learning control pipeline. We validate the approach in simulation and on a Unitree G1 humanoid, showing robust sim-to-real transfer and consistent audio-conditioned policy selection. Supplementary materials are available at the following site: \url{https://lab-rococo-sapienza.github.io/semantic-WBC/}

\end{abstract}
\section{Introduction}

The landscape of humanoid robotics has seen an impressive acceleration in recent years. Although the field has been active for decades, the joint progress of humanoid robots capable of moving in a very natural way and powerful AI models has transformed humanoid platforms from experimental laboratory prototypes into systems increasingly present in the social fabric.

The growth is also shown in international competition and scenario such as the scenario of RoboCup.
One of the first changes that RoboCup has seen is the merging of Humanoid and Standard Platform Leagues into a unified Humanoid Soccer League (HSL), providing a single competitive stage for humanoid robots of diverse dimensions to demonstrate their capabilities in playing soccer autonomously in the dynamic scenarios of the soccer matches.

The evolution of machine learning models, specifically reinforcement learning algorithms  moved the paradigm of the robotic locomotion and manipulation \cite{radosavovic2024real,hwangbo2019learning}, and nowadays modern humanoids exhibit fluid movements that closely approximate human movements \cite{cheng2024expressive}, moving beyond static, predefined paths toward dynamic stability and high-level agility.

\begin{figure}[t]
    \centering
    \includegraphics[width=0.8\linewidth]{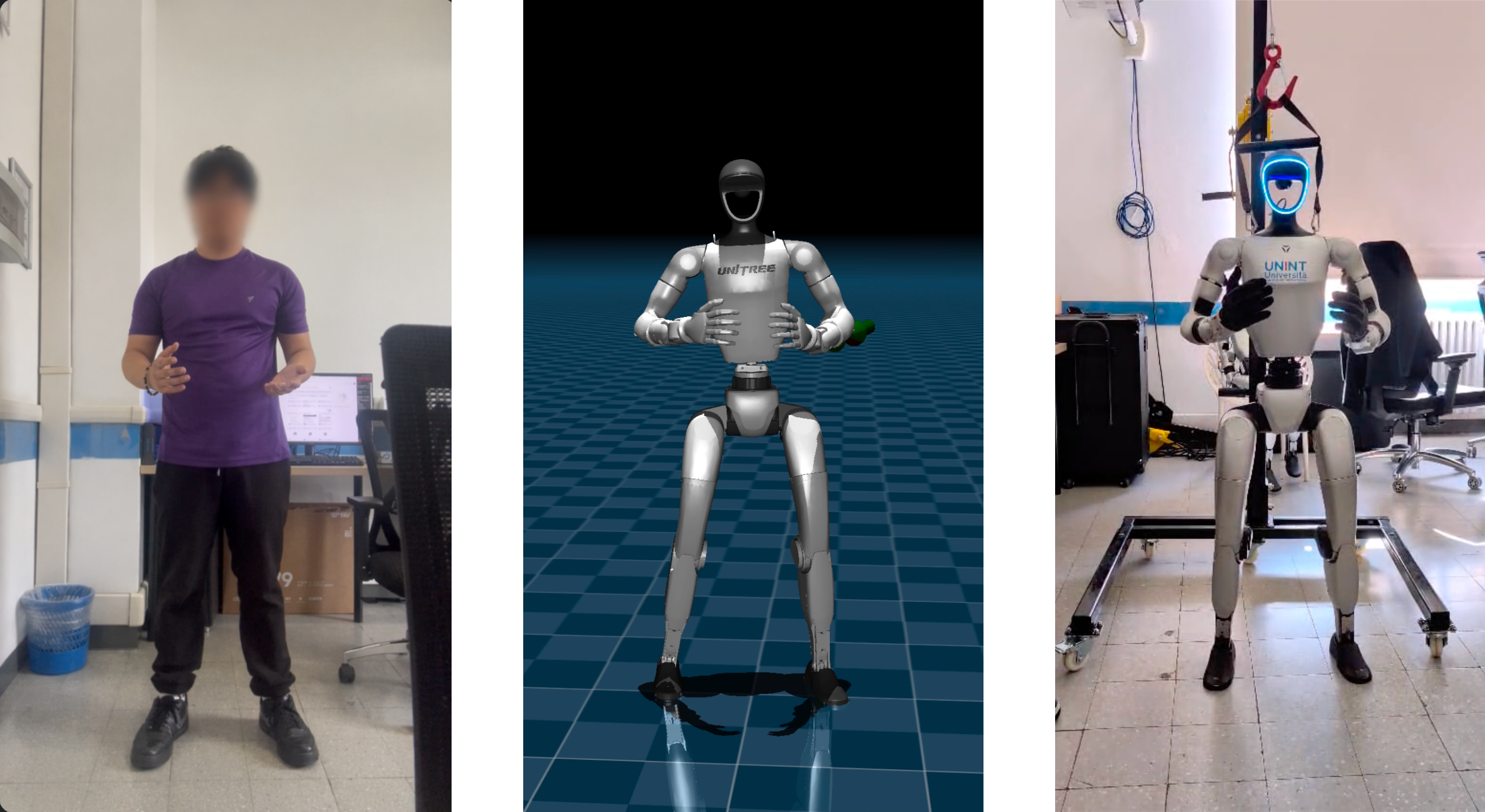}
    \caption{Example of human motion retargeting in simulation and real-world deployment on the Unitree G1 robot.}
    \label{fig:introduction}
\end{figure}

Such evolution in control has paved the way for humanoid robots to engage in artistic expression. This is already evident in large-scale entertainment events, particularly in China, where robots perform synchronized dances at concerts or social events \cite{cnnchina}. Such examples demonstrate the evolution of robots from laboratory prototypes to autonomous agents usable in public show collaborating with humans in a natural way. The mimicking of complex human rhythmic and spatial patterns is creating a new frontier for live entertainment, the impact of which on the audience's perception is generated by the ability to perform human-like actions as shown in Fig. \ref{fig:introduction}.

RoboCup ecosystem also focus on the intersection of robotics and artistic expression. An example is the RoboCupJunior OnStage competition, which challenges teams to design and program autonomous physical robots to deliver live performances such as dance or specific choreography that engage a human audience. Success in this league requires the innovative integration of sensory technologies to enhance the visual and entertainment value of the choreography.

However, in the vast majority of cases, robotic performances still rely on time-coded sequences and fixed rules \cite{infantino2016robodanza}, limiting their ability to adapt to dynamic contexts. Recent advances in Reinforcement Learning (RL) and imitation learning have made it possible to learn highly complex humanoid motions. Despite this progress, these behaviors are typically not context-aware: a robot may execute a movement accurately, but it lacks an intrinsic understanding of when and why that action should be performed.

This limitation highlights the need for a mechanism that connects perception and action. A promising direction is to introduce embedding spaces that link physical actions to semantic meaning, enabling the robot to associate movements with their contextual relevance and to interpret “what” it is doing in relation to its environment. Such an approach replaces rigid rule-based triggers with a perception-driven selection mechanism, allowing the robot to autonomously choose appropriate behaviors at runtime while leveraging existing motion-learning techniques.

We present a multi-modal orchestrator framework for autonomous action selection via real-time sensor integration. The system processes audio streams to infer the current context and trigger the appropriate motion policy, while natural language commands provide additional flexibility. This unified interface keeps low-level motor skills semantically grounded.
By mapping auditory cues to a library of imitation-learned motion policies, the robot dynamically adapts its whole-body movements to the perceived environment, avoiding the mechanical appearance of rule-based systems. This enables the robot to effectively "listen," "understand," and respond in real time, supporting engaging autonomous scenarios such as RoboCup OnStage.
The main contributions are threefold:
\begin{itemize}
    \item We propose a unified retrieval module that maps short audio chunks to the most relevant motion policy (top-1), supporting both \emph{music-based} alignment (via audio fingerprinting and embeddings) and \emph{text-based} grounding (via speech transcription and semantic matching). This enables real-time selection of appropriate behaviors directly from perceptual input.
    \item We design and implement a complete end-to-end system that integrates streaming audio processing, hierarchical routing, retrieval, and execution over a library of imitation-learned humanoid skills. The pipeline is modular, and compatible with both simulation and real-world deployment.
    \item We show that embedding semantic information into the control loop enables context-aware behavior selection, moving beyond pre-scripted choreography.
\end{itemize}
The remainder is organized as follow.  Section \ref{sec:rel_work} reviews related work, Section \ref{sec:methodology} presents the proposed approach from the retrieval to the skill execution, Section \ref{sec:results} reports the experimental results, and Section \ref{sec:conclusion} concludes the paper.

\section{Related Works}
\label{sec:rel_work}

Recent advances in reinforcement learning and imitation learning have greatly expanded the expressive capabilities of humanoid robots, enabling increasingly natural and complex whole-body motions \cite{radosavovic2024real,radosavovic2024learning,liao2025beyondmimic}. 
Learning from human demonstrations has emerged as a scalable route toward agile humanoid behavior, avoiding hand-designed reward functions for complex stylistic motions. DeepMimic \cite{peng2018deepmimic} showed that standard RL algorithms can robustly imitate motion-capture clips, while AMP \cite{peng2021amp} removed handcrafted imitation objectives by learning style rewards from unstructured motion datasets. ExBody \cite{cheng2024exbody}, H2O and OmniH2O \cite{he2024learning,he2024omnih2o}, and HumanPlus \cite{fu2024humanplus} progressively expanded the scope of imitable behaviors, from expressive upper-body motion to multi-modal teleoperation and autonomous whole-body skills. 

A key prerequisite in these systems is \emph{retargeting}, i.e., mapping human motion onto a humanoid body with different proportions, degrees of freedom, and joint limits. TWIST \cite{ze2025twist} combines kinematic retargeting with whole-body control trained through reinforcement learning and behavior cloning, enabling coordinated locomotion and manipulation. General Motion Retargeting \cite{araujo2025retargeting} shows that retargeting quality itself is a major bottleneck. Vision-based methods such as WHAM \cite{shen2024world} and faster 3D perception backbones \cite{yang2026fast} further broaden the range of usable demonstration sources by extracting skeletal motion directly videos. Collectively, these systems demonstrate that a humanoid can acquire a rich library of expressive skills but the question of how to select among them autonomously from continuous sensory input represent a problem statement.
One of the example where to stress-test this capability is dance since is one of the discipline that requires motor coordination, temporal synchronization with music, and orchestration of stylistically diverse motions. On the generation side, AI Choreographer \cite{li2021ai} proposed long-sequence 3D dance synthesis conditioned on music, Bailando \cite{siyao2022bailando} formulates music-to-dance generation as a choreographic memory indexed by an actor-critic GPT, and EDGE \cite{tseng2023edge} uses a transformer-based diffusion model with Jukebox audio embeddings and joint-wise conditioning for editable dance generation. These methods produce high-quality kinematic sequences but do not close the loop with a physical robot. 

On the execution side, RobotDancing \cite{sun2025robotdancing} transfers retargeted human dance onto real humanoids through residual-action reinforcement learning, improving long-horizon tracking, while MDRPC \cite{guan2024mdrpc} learns reusable dance primitives through adversarial imitation and organizes them according to musical structure, and DanceHAT \cite{nie2022dancehat} reduces reliance on hand-crafted robot-specific features through adversarial training. RoboPerform \cite{li2026roboperform} further proposes an audio-conditioned framework for music-driven dance and co-speech gestures, establishing audio as a powerful control modality for expressive robotic behavior. A common thread across all these systems is that the orchestration of motion what to perform and when is determined through an offline pipeline, rather than inferred online and continuously from live sensory input without explicit human intervention.
Parallel work has explored language as an interface for high-level skill orchestration. OmniH2O \cite{he2024omnih2o} integrates GPT-4 as an autonomy driver over a whole-body teleoperation policy, while FRoM-W1 \cite{li2026w1} proposes an open-source framework for language-conditioned humanoid control. In robotic manipulation, VLM-driven skill selection over pre-trained primitive libraries \cite{ingelhag2024robotic,gao2024physically} demonstrates that semantic reasoning can effectively bridge perception and low-level control. These systems establish the feasibility of high-level semantic orchestration, but rely on discrete text-based commands as the control interface a modality that requires explicit human intervention at each decision point.
\section{Methodology}
\label{sec:methodology}

\begin{figure}[t]
\centering
\includegraphics[width=\linewidth]{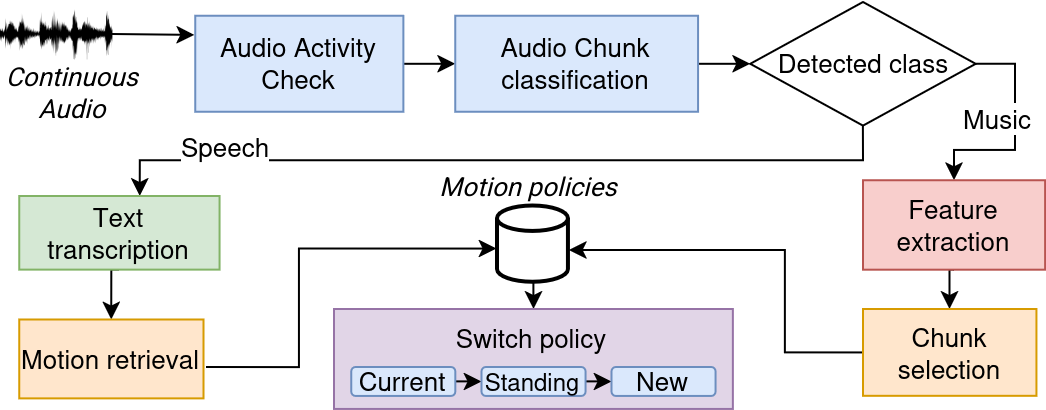}
\caption{\textbf{Proposed architecture}. Continuous microphone audio is segmented into fixed-length chunks, each chunk is assigned a skill identifier by a routing and retrieval chain, and the chosen skill is executed on the humanoid through a switching layer that mediates transitions. }
\label{fig:architeceture} 
\end{figure}

The proposed method shown in Fig. \ref{fig:architeceture} converts a streaming audio signal into online decisions over a fixed subset of motion policies. For each chunk $x_k$, the pipeline performs three operations in sequence. First, it classifies the perceived chunk into a music, speech, or non-informative class. Second, it retrieves track identity and temporal offset for musical chunks, or grounds linguistic intent into a skill identifier for transcribed audio chunks. Finally, it dispatches at most one command that either switches the active skill or triggers a time-boxed gesture.
The listener records in a continuous way the microphone input and segments it into fixed-length chunks of fixed duration $T$ (we choose 5 s for our tests). Each chunk is given in input to the pipeline for the retrieval.
To avoid spurious skill switches, the system enforces a set of acceptance conditions before any command is emitted. On the music side, a retrieved match is accepted only if both its confidence score and the number of aligned landmark votes exceed minimum thresholds, discarding matches that are statistically weak or poorly supported. On the speech side, a chunk is forwarded to the transcription model only if the detected voice activity is sufficiently strong and clearly dominant over any residual musical content, preventing ambiguous mixed-content segments from triggering unintended commands. Finally, a cool-down mechanism suppresses repeated emissions of the same skill identifier within a short time window, damping oscillations that may arise at section boundaries where the audio content transitions gradually between two tracks.
 
\subsubsection{Hierarchical audio routing}
\label{sec:methodology:routing}

Each audio chunk is classified into one of three mutually exclusive branches: \textsc{Music}, for segments containing musical content; \textsc{Speech}, for segments carrying vocal content; and \textsc{Skip}, for not relevant chunks. The routing decision combines two complementary signals. First, the \emph{Audio Spectrogram Transformer} (AST)~\cite{Gong2021ASTAS}, a convolution-free Transformer that operates on log-Mel spectrogram patches, is employed as a pre-trained classifier over the 527-class AudioSet ontology~\cite{Gemmeke2017AudioSA}. Its output posteriors are aggregated into two scene-level scores, $p_\text{music}$ and $p_\text{speech}$, by summing the probabilities of the respective label groups. Second, Silero VAD~\cite{SileroVAD}, a lightweight neural voice-activity detector, is applied to the 16~kHz waveform to produce per-frame speech probabilities; the fraction of frames exceeding the detection threshold, $v_\text{frac}\!\in\![0,1]$, serves as a voicing indicator.
The router assigns a chunk to \textsc{Speech} if $v_\text{frac} \ge v^{\min}_\text{frac}$ and $p_\text{speech} - p_\text{music} \ge \epsilon_\text{sp}$, to \textsc{Music} if $p_\text{music}$ dominates and $v_\text{frac}$ remains below threshold, and to \textsc{Skip} otherwise.
 
\subsubsection{Music retrieval with temporal skill grounding}
\label{sec:methodology:music}
 
Chunks in the \textsc{music} branch are processed by a local audio-fingerprinting module that follows the Wang \emph{constellation-map} algorithm~\cite{Wang2003a}: each reference track is indexed by hashing pairs of spectrogram peaks together with their relative time offset, and a query chunk is matched by retrieving candidate pairs and looking for a dominant peak in the histogram of aligned temporal offsets. The index is serialised at build time from the catalogue of reference tracks. For a chunk $x_k$, the matcher returns a tuple $(s, c, v, \tau)$, where $s$ denotes the candidate track identifier, $c\!\in\![0,1]$ the matching confidence derived from the vote distribution, $v$ the number of aligned landmark votes, and $\tau$ the estimated offset of the chunk within the reference track. When the fingerprint match falls below the acceptance thresholds, the branch falls back to a cosine search in the embedding space of the Contrastive Language-Audio Pretraining (CLAP) model~\cite{CLAP}, a dual audio-text encoder trained contrastively to map short audio segments into a joint semantic space. 
 
Retrieval is grounded temporally: the offset $\tau$ selects among different skills within the same track. We define a set of \emph{timed skill rules}
\[
r = \bigl(s^{\star},\; [t_\text{start},\; t_\text{end}),\; \pi\bigr)
  \in \mathcal{R},
\]
parsed from text descriptors. Given an accepted hit $(s, c, v, \tau)$, the resolver returns the skill identifier~$\pi$ of the first rule $r \in \mathcal{R}$ with $s = s^{\star}$ and $\tau \in [t_\text{start}, t_\text{end})$. If no timed rule applies, the resolver falls back to a track-level mapping $s \mapsto \pi$. Successive passages of a single track (intro, verse, chorus, outro) therefore drive distinct whole-body behaviours without retraining the underlying controllers.
\subsubsection{Speech-to-motion grounding}
\label{sec:methodology:speech}

\begin{figure}[t]
    \centering
    \includegraphics[width=0.8\linewidth]{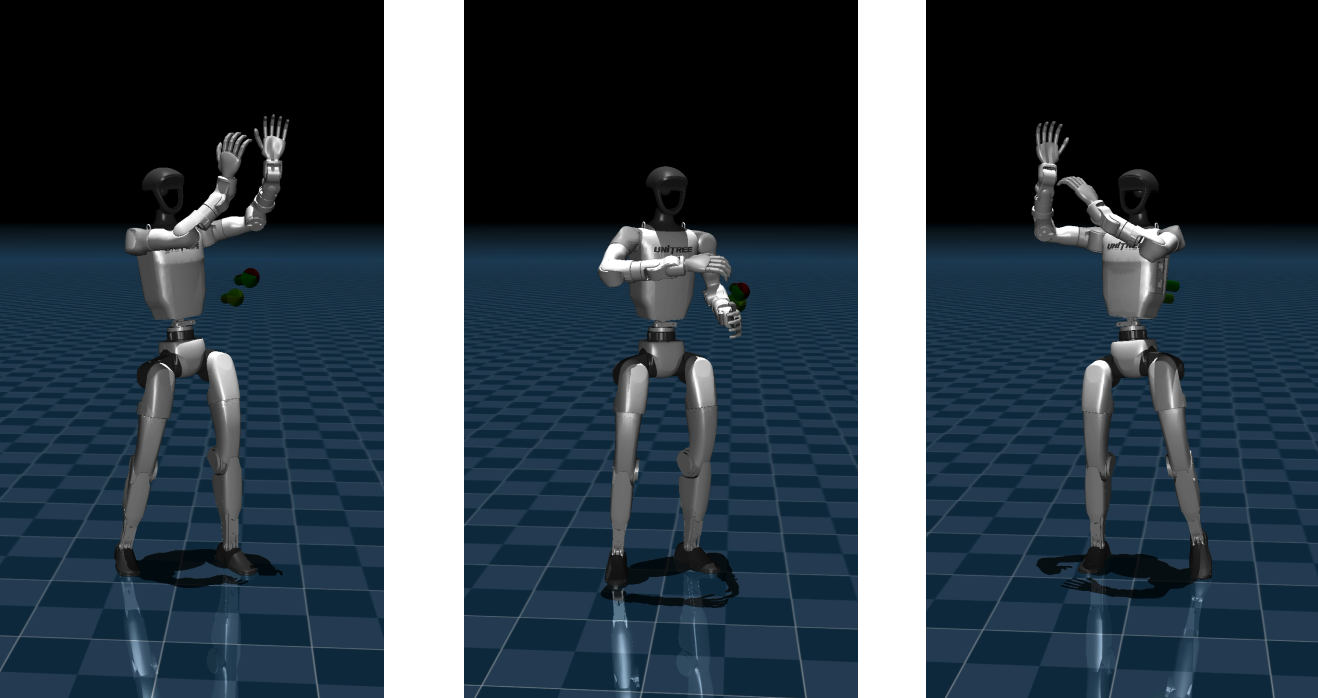}
    \caption{Sim-to-sim execution in the RoboJuDo~\cite{RoboJuDo} environment.}
    \label{fig:sim_dance}
\end{figure}

Chunks in the \textsc{speech} branch are transcribed by a streaming speech-to-text model. In our implementation we use the OpenAI \texttt{gpt-4o-mini-transcribe} endpoint, a low-latency variant of Whisper-class autoregressive transcription models. The transcript is matched against the skill library to retrieve the top-1 motion policy via semantic grounding. In parallel, the system supports direct human-robot interaction: unmatched transcripts are forwarded to a conversational LLM (\texttt{gpt-4o-mini}), whose response is synthesised by a neural text-to-speech model and played through the robot speaker. The estimated speech duration is simultaneously transmitted to the robot, triggering a fixed gesture policy of matching duration and resulting in natural and temporally coherent full-body communicative behaviour.
 
\subsubsection{Command interface and skill execution}
\label{sec:methodology:runtime}

Both branches communicate with the robot through a unified TCP socket interface implemented in RoboJudo\cite{RoboJuDo}, our control framework for orchestrating multi-policy humanoid behaviors. The music and speech modules emit skill identifiers that are queued and executed by the low-level control pipeline. The skill library consists of two base policies, a locomotion (Walk) policy and a Stand policy used to mediate safe transitions, alongside a configurable set of imitation-learned whole-body policies trained with BeyondMimic~\cite{liao2025beyondmimic}, a reinforcement learning framework that imitates reference human-motion clips and exports the resulting controllers as ONNX policies for real-time inference. When a new skill is requested, the system optionally primes the robot through the Stand policy before activating the target motion, ensuring stable transitions between heterogeneous whole-body behaviors. The control loop runs inside a MuJoCo~\cite{todorov2012mujoco} simulation of the Unitree~G1 humanoid and is deployed on the physical platform by substituting the simulated environment with the Unitree hardware interface, with no changes to the orchestration logic.

\section{Experimental Results}
\label{sec:results}

In the experimental evaluation we evaluated whether the selected policies produce a coherent choreography when executed in sequence, using two mashup scenarios of different temporal granularity. Starting from the songs we split them into fixed 5-second chunks. To evaluate retrieval not only under perfectly aligned conditions, but also under more realistic listening conditions in which the system may start receiving audio with a small delay after the song has already begun, we additionally generated shifted chunks by introducing an initial temporal offset of 0.5, 1.0, 1.5, and 2.0 seconds. This allows us to test whether the system can still retrieve the correct policy even when the observed audio segment does not exactly coincide with the original chunk boundaries used in the embedding index. A chunk was counted as correctly classified if the policy retrieved by the system matched the ground-truth policy associated with that specific audio fragment. Across a total of 574 evaluated chunks under these conditions, the method achieved an overall chunk-level accuracy of 84.8\%.

In addition to the quantitative benchmark in simulation, we also include a qualitative real-world validation on the Unitree G1 platform to verify that the same high-level control logic transfers outside the offline setting.

\begin{figure}[t]
    \centering
    \includegraphics[width=1\linewidth]{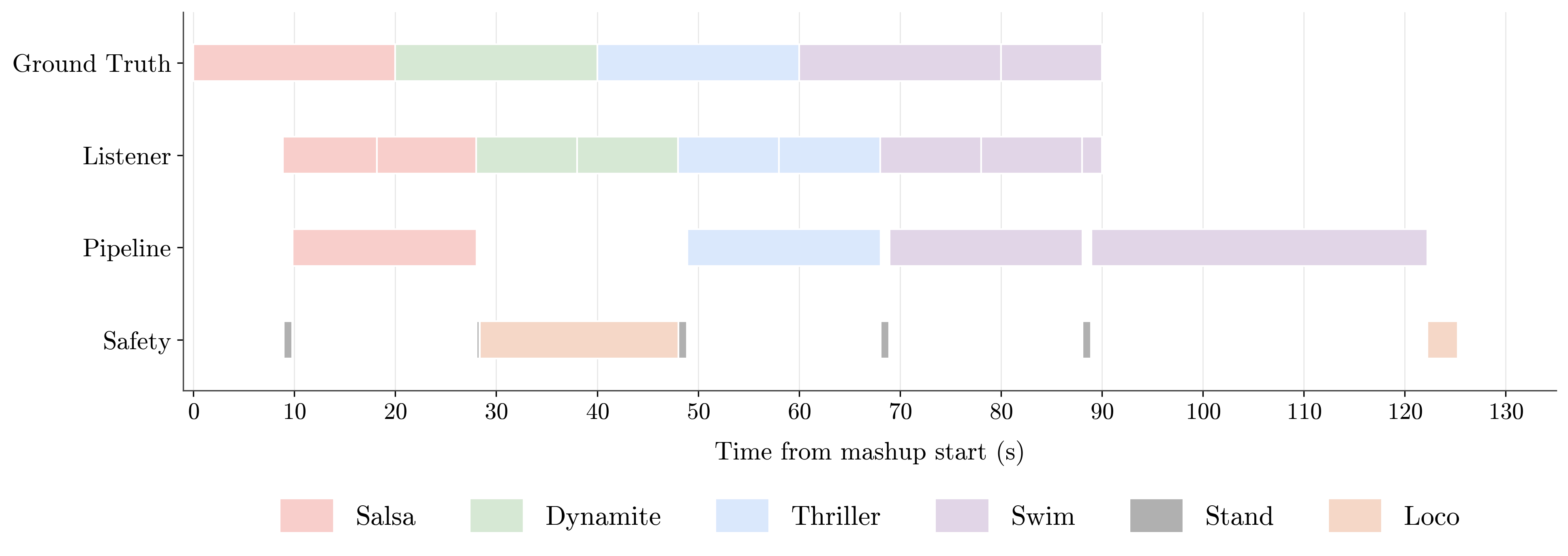}
    \caption{Gantt chart for the 20-second mashup (M20). The controller struggles 
    to complete transitions within the available time window, eventually falling 
    back to locomotion mode due to a stability trigger.}
    \label{fig:20sec}
\end{figure}

\begin{figure}[t]
    \centering
    \includegraphics[width=1\linewidth]{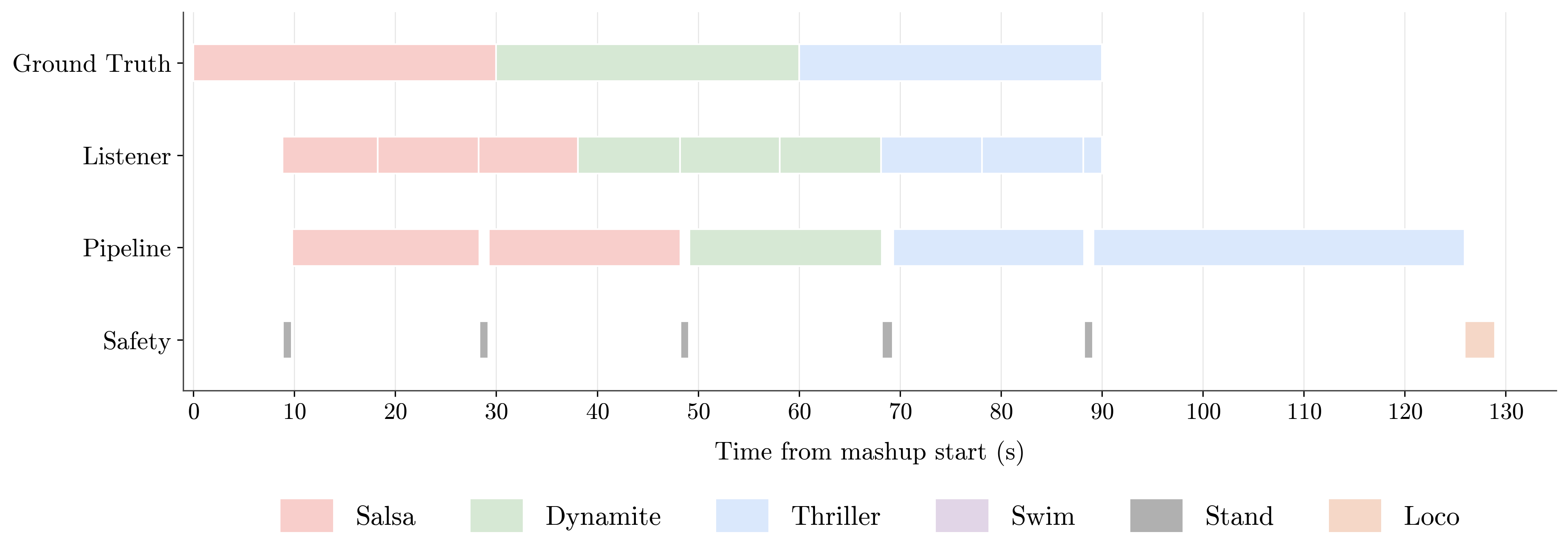}
    \caption{Gantt chart for the 30-second mashup (M30). Policy transitions remain 
    well aligned with the commanded sequence for most of the execution, with only 
    a slight prolongation in the final phase.}
    \label{fig:30sec}
\end{figure}
\subsection{Choreography Execution in Simulation}
\label{sec:evaluation:simulation}
Given the retrieval accuracy reported above, we further evaluate whether the selected policies produce a coherent choreography when executed in sequence. Two mashup scenarios are constructed in simulation, where the musical track changes every 20 seconds (M20) and every 30 seconds (M30) respectively. Each mashup consists of four musically distinct tracks: \textit{Salsa}, a Latin dance piece; \textit{Dynamite} and \textit{Swim}, two K-pop songs by BTS; \textit{Thriller}, a pop track by Michael Jackson. The execution is visualized through Gantt charts that compare the commanded policy sequence with the one actually applied by the controller.

\begin{figure}[t]
    \centering
    \includegraphics[width=1\linewidth]{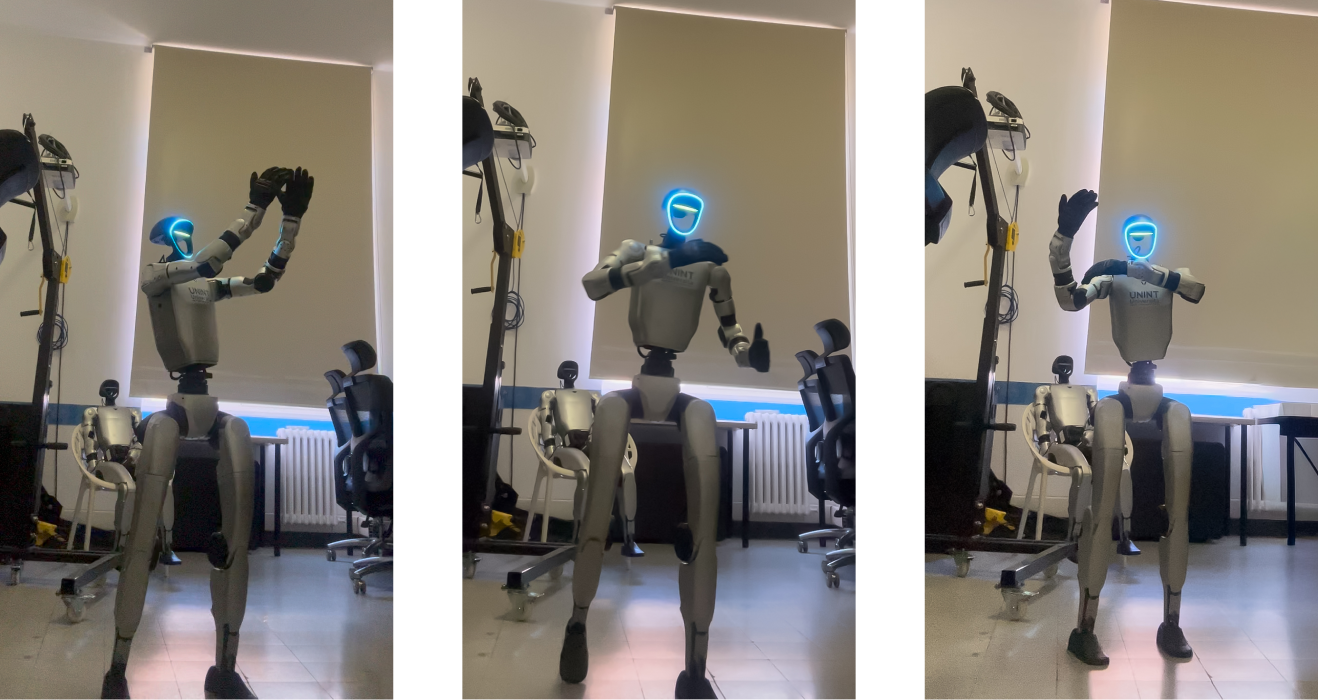}
    \caption{Real-world execution on the Unitree G1, where music recognition 
    drives the motion policy in a live setting.}
    \label{fig:real_dance}
\end{figure}

As shown in Figures~\ref{fig:20sec} and~\ref{fig:30sec}, the 30-second setting produces a significantly more consistent execution. In M30, the controller has sufficient time to complete each transition through the intermediate stand phase before the next policy is requested, resulting in a choreography that closely follows the commanded sequence. In M20, the shorter sections do not leave enough time for transitions to stabilize, causing misalignments and occasional fallbacks to the locomotion policy;  in the most critical cases, the transition drives the robot toward a loss-of-balance condition, effectively pushing the center of mass outside the support polygon and triggering the safety mechanism. These results indicate that 30-second musical sections represent a suitable operating point for the current transition mechanism, and that retrieval quality is not the primary bottleneck: the main limiting factor is the transition latency introduced by the stand-priming phase.

\subsection{Real-World Validation}
\label{sec:evaluation:realworld}

Building on the simulation results, the same framework is deployed on the physical Unitree G1 humanoid to verify that the control logic transfers to a real execution setting. The real-world demonstration confirms that chunk-level retrieval is sufficiently stable to drive consistent policy selection in a live scenario. As expected, end-to-end latency is higher than in simulation due to network-dependent calls to external providers, the communication overhead of the hardware interface, and the current policy-switching mechanism, which routes transitions through an intermediate standing state for stability. Despite this, the robot successfully follows the commanded policy sequence, demonstrating the practical applicability of the approach.

\section{Conclusion}
\label{sec:conclusion}
In this work, we presented a multi-modal orchestration framework for semantic audio-driven humanoid control, enabling robots to autonomously select and execute whole-body skills from continuous audio streams. By integrating hierarchical audio routing, temporal grounding for music, and speech-based skill selection, the system bridges the gap between perception and action through a unified policy scheduling interface.
A key advantage of this approach is its generalization capability: by leveraging embedding spaces, the framework can map novel audio inputs to the most semantically similar motion policies without requiring additional policy training. While the scalability of this method warrants further validation through extensive testing, the current results demonstrate significant potential beyond dance choreography. This architecture opens the way for diverse robotic applications where complex motion can be effectively grounded in cross-modal embeddings.
Furthermore, experimental results show that our approach supports robust, context-aware behavior selection both in simulation and on the real Unitree G1 platform. This demonstrates the feasibility of moving beyond pre-scripted robotic performances toward more adaptive, responsive, and autonomous behaviors in dynamic environments. Future work will extend the framework with a style-change detection module that enables adaptive audio segmentation based on musical pattern analysis, reducing the latency introduced by fixed-length chunking and improving temporal alignment between motion and music.

\begin{credits}
\subsubsection{\ackname} Michele Brienza is funded by the European Union - Next Generation EU, Mission I.4.1 Borse PNRR Pubblica Amministrazione (Missione 4) Component 1 CUP B53C23003540006.
\end{credits}

\bibliographystyle{splncs04}
\bibliography{biblio}
\end{document}